\documentclass{article}
\usepackage{arxiv}
\pdfoutput=1

\usepackage[utf8]{inputenc} 
\usepackage[T1]{fontenc}    
\usepackage{hyperref}       
\usepackage{url}            
\usepackage{booktabs}       
\usepackage{amsfonts}       
\usepackage{nicefrac}       
\usepackage{microtype}      
\usepackage{lipsum}
\usepackage{graphicx}
\graphicspath{ {./images/} }
\usepackage{subfigure}

\title{Leveraging Pre-trained and Transformer-derived Embeddings from EHRs to Characterize Heterogeneity Across Alzheimer's Disease and Related Dementias}

\author{
 Matthew West\textsuperscript{1,2}, Colin Magdamo\textsuperscript{2}, Lily Cheng\textsuperscript{2}, Yingnan He\textsuperscript{2}, Sudeshna Das\textsuperscript{2}\\
  \textsuperscript{1}Department of Biostatistics, Harvard T.H. Chan School of Public Health, Boston, MA \\
    \textsuperscript{2}Department of Neurology, Massachusetts General Hospital and Harvard Medical School, Boston, MA \\
    Correspondence: \texttt{sdas5@mgh.harvard.edu} \\
}

\begin{document}
\maketitle
\begin{abstract}
Alzheimer's disease is a progressive, debilitating neurodegenerative disease that affects 50 million people globally. Despite this substantial health burden, available treatments for the disease are limited and its fundamental causes remain poorly understood. Previous work has suggested the existence of clinically-meaningful sub-types, which it is suggested may correspond to distinct etiologies, disease courses, and ultimately appropriate treatments. Here, we use unsupervised learning techniques on electronic health records (EHRs) from a cohort of memory disorder patients to characterise heterogeneity in this disease population. Pre-trained embeddings for medical codes as well as transformer-derived Clinical BERT embeddings of free text are used to encode patient EHRs. We identify the existence of sub-populations on the basis of comorbidities and shared textual features, and discuss their clinical significance.
\end{abstract}


\section{Introduction}

Alzheimer's disease (AD) is a degenerative neurological condition which affects around 50 million people globally, and is the 6th leading cause of death in the US \cite{mebane20092009}. Despite its substantial public health burden, AD remains poorly understood and there are relatively few interventions or treatment options available. AD is a type of dementia, a syndrome that emerges following neurodegeneration, and typically affects memory, cognition and behaviour. Alzheimer's disease and related dementias (ADRD) is an umbrella term that refers to AD and a number of associated common dementias, including frontotemporal dementia (FTD), Lewy body dementia (LBD), and vascular dementia. AD is the most common among the ADRD pathologies, and constitutes around 60-80\% of all dementias, being distinguished from other dementias by exerting more significant impact on memory \cite{mebane20092009}.

Pathophysiologically, AD is characterised by the build-up of amyloid-$\beta$ plaques and neurofibrillary tangles constituted by hyperphosphorylated tau proteins. The so-called `amyloid hypothesis' has prevailed as the leading explanation of disease etiology, where it is held that amyloid-$\beta$ toxicity leads to synaptic dysfunction and neurodegeneration \cite{hardy2002amyloid}. However, while most treatments in development have aimed to target this amyloid hypothesis, existing treatments exhibit borderline efficacy or only show benefit limited to small subpopulations. This motivates an approach to disease characterisation where potential subtypes of disease could be elucidated, thereby informing different disease etiologies, trial designs, and ultimately the development of effective treatments.

Previous approaches to subtyping Alzheimer's disease have focused on molecular subtyping on the basis of RNA expression, as well as subtyping and staging on the basis of brain imaging and cognitive assessments. Neff \textit{et al.} \cite{neff2021molecular} use RNA-seq signatures from the brain to identify five molecular subtypes of AD in three major classes, characterised by different combinations of multiple dysregulated pathways. These broadly pertain to tau-mediated neurodegeneration, amyloid-$\beta$ neuroinflammation, synaptic signaling, immune activity, mitochondria organization, and myelination.

Using imaging data from patients with FTD and AD, the SuStaIn (Subtype and Stage Inference) algorithm is able to extract disease stage and subtype by training a probabilistic generative model of disease progression on time-series of MRI images \cite{young2018uncovering}. From this, three distinct trajectories of AD were identified, characterised anatomically by the rate and sequence of atrophy in different brain regions. The SuStaIn algorithm was later applied to a multi-site cohort of positron emission tomography (PET) scans, revealing distinct trajectories of tau deposition in AD \cite{vogel2021four}. 

There has also been prior work characterising subtypes of disease based on neuropsychological profiles assessed from cognitive tests. These broadly group patients on the basis of memory, visuospatial and linguistic capabilities, and executive function \cite{libon2014neuropsychological, scheltens2016identification, scheltens2017cognitive}. Aside from molecular, imaging, and cognitive data, some prior work clustering AD has utilised electronic health records (EHRs) as a modality. EHRs offer easy access to large observational datasets for clinical research, but are subject to their own inherent biases due to not being designed for research purposes. Unsupervised learning on EHR datasets has been leveraged successfully to reveal latent structure in other neurological and neurodevelopmental conditions, such as autism  \cite{doshi2014comorbidity, luo2020multidimensional}, as well as Parkinson's disease \cite{zhang2019data}.

In the context of Alzheimer's disease, these EHR-based approaches have been applied with some success, with varying results dependent on methodology and population. Xu et al. \cite{xu2020data} train an AD risk-prediction model on an EHR-derived cohort, using structured features like diseases and medications to predict EHR-derived AD status. They then cluster representations for AD patients derived from predictive features in the risk-prediction task, identifying  subtypes by prevalence of comorbidities and medications. These subtypes broadly correspond to patients with cardiovascular disease, those with mental illness, those with older age of onset, and finally those taking medications for dementia and with sensory problems.

Alexander \textit{et al.} \cite{alexander2020using} use a large EHR system to characterise the clinical heterogeneity of AD, using diagnoses, symptoms, and demographic data from a literature-informed feature set to characterize patients. They provide a comparative analysis of four different clustering methods and assess the stability and reproducibility of the phenotypic clusters that are identified. They reliably identify a subpopulation of early-onset AD patients characterised by a faster rate of progression and being predominantly female. 
The approach in Landi \textit{et al.} \cite{landi2020deep} utilises unsupervised deep learning to derive representations of EHRs from an entire hospital system, without restriction to any one disease cohort. To derive these representations, they explicitly encode each EHR in a temporal sequence, constituted by diagnoses, medications, labs, and procedures along with pre-trained word embeddings to characterise medical concepts extracted from notes. These representations are passed through a convolutional autoencoder, and latent representations are retained to cluster patients. Following training, they cluster on patient representations to try and identify separation between 8 disease groups, before attempting to characterize heterogeneity within disease group. One of these disease groups is AD, where they identify three distinct clusters: patients with early-onset disease, patients with later-onset disease and mild neuropsychiatric and cerebrovascular comorbidities, and finally patients with typical onset and mild-to-moderate dementia symptoms.

More recently, He \textit{et al.} \cite{he2022temporal} performed a spectral clustering on a cohort of patients EHRs, using features derived in a 3-year window prior to AD onset. They specifically encode diagnoses prior to AD onset in discrete 6-month intervals, and identify four clusters that differ significantly in terms of demographics, mortality, and prescription medications. Tang \textit{et al.} \cite{tang2022deep} perform a deep clinical phenotyping on the basis of comorbidities in patient EHRs from two independent cohorts, and identify a number of clusters that appear to exhibit dependence on sex. For example, after sex stratification there is an enrichment for vascular and musculoskeletal disorders in females with AD vs female controls, and enrichment for behavioural/neurological disorders in males with AD vs male controls. They also characterise network interactions between comorbidities in AD patients, which compared to controls exhibit greater numbers of edges and higher rates of comorbid conditions. 

Building on these studies, the methodology contained in this work offers the following contributions:

\begin{itemize}
    \item We make use of pre-trained embeddings from a medical context to encode ICD codes prior to clustering.
    \item In contrast to prior work, we make direct use of clinical free-text without prior rule-based extraction of medical concepts.
    \item We demonstrate the use of transformers adapted to a clinical domain to encode representations of clinical text, and that these representations are able to characterise heterogeneity in AD following clustering.
\end{itemize}

In this work, we both utilise both pre-trained embeddings for ICD code diagnostic data and transformer-derived embeddings to characterise clinical notes, clustering patient representations for each of these modalities independently. Transformers are a recent development in deep learning that have exhibited state-of-the-art performance in language modelling tasks, and are broadly characterised by their reliance on the concept of attention \cite{vaswani2017attention}. Attention, so-called because it mimics cognitive attention, allows contextual information to be shared between word representations without explicitly encoding word sequence, by providing attention weights to bias important parts of prior representations in the network's forward pass. The transformer used in this work is a version of the BERT architecture fine-tuned on a biomedical corpus \cite{devlin2018bert}. BERT allows for improved contextual representations over the original transformer architecture, with its base architecture having 12 attention heads in each of its 12 layers, allowing for more detailed contextual word embeddings. This architecture provides an encoder which can be fine-tuned on downstream applications and domains, as done in BioBERT, which fine-tunes the BERT-base architecture on PubMed abstracts and PMC full-text articles  \cite{lee2020biobert}. Specifically, we use Clinical BERT, a publicly-released model which further fine-tunes BioBERT on notes from the critical care database MIMIC  \cite{alsentzer2019publicly, johnson2016mimic}.

\section{Methods}

\subsection{Cohort Selection Process}

\begin{figure}
    \centering
    \includegraphics[width=0.9\textwidth]{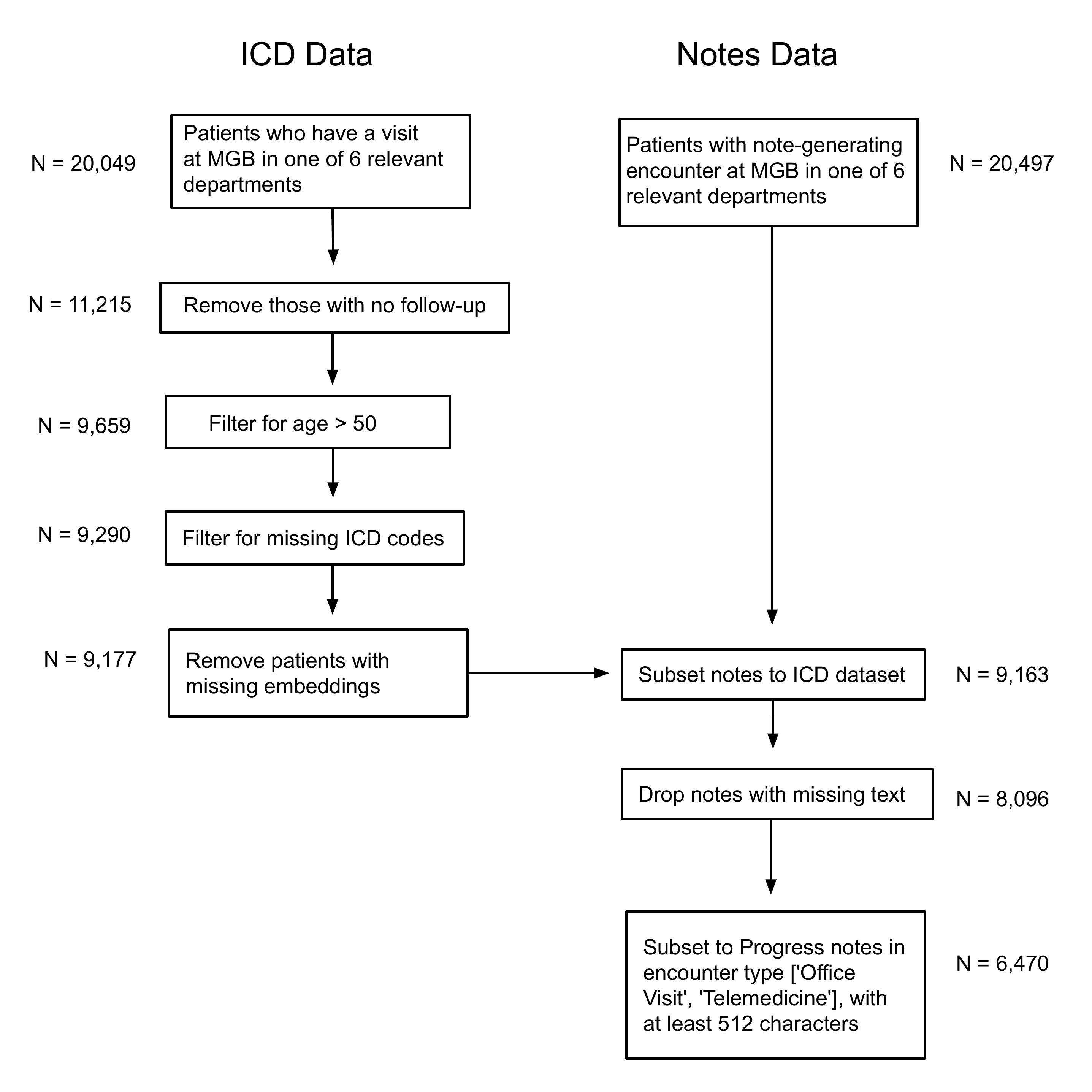}
    \caption{\textbf{Cohort selection process.} A flowchart showing the cohort selection process for the our patient cohort.}
    \label{fig:cohort}
\end{figure}

Patients were queried from the Mass General Brigham (MGB) Research Patient Data Registry (RPDR) query tool, where $N=20,497$ candidate patients having at least one note-generating encounter in one of 6 relevant departments having memory specialists were identified. Patients were drawn mainly from the Memory Disorders Unit at Massachusetts General Hospital (MGH), as well as patients seen by memory specialists at Brigham and Women's Hospital (BWH). Two datasets were extracted in parallel for the patient cohort identified, one of structured diagnostic ICD code data, as well as one of unstructured clinical notes written by a neurologist. Access to this data for research purposes was approved by the Mass General Brigham Institutional Review Board (IRB).

$N=20,049$ patients were identified from memory disorder department visits on the basis of ICD codes alone. Subsetting to those with at least one follow-up encounter, having an age at first encounter greater than 50 years, and filtering for missing ICD9 codes yielded $N=9,290$. After converting patient records to vector representations using pre-trained embeddings, patients with missing embeddings were dropped to yield an ICD dataset of $N=9,177$ patients. After dropping patients without notes from a neurologist encounter, and selecting for only substantial notes ($\geq 512$ characters) documenting office visits or telemedicine encounters, the final cohort was constituted by $N=6,470$ patients. This cohort selection process is shown in more detail in Fig. \ref{fig:cohort}.

For characterising our ADRD phenotype in the cohort and in our resultant clusters, the presence of at least one relevant ICD9 code for each disease group was used. For AD this was 331.0, and for FTD this was 331.1 (FTD), 331.19 (other FTD) and Pick's disease (331.11). For LBD, 331.82 was used, and vascular dementia was defined by having any of 290.40 (uncomplicated), 290.41 (with delirium), 290.42 (with delusions), or 290.43 (with depressed mood). Finally, MCI was defined by the presence of 331.83, and memory loss by 780.93.

\subsection{Embedding Methodology}
\subsubsection{ICD Codes}
Prior to clustering, it was necessary to derive a patient-level representation that encodes information relevant to phenotype in a single vector. While some prior work has relied on one-hot encoding of clinical data to represent patient phenotype, we leverage existing pre-trained embeddings that capture relevant biomedical semantics in their latent representations of clinical concepts. In particular, we use a set of 300-dimensional embeddings for ICD9 codes, derived from prior work by Choi \textit{et al.} \cite{choi2016learning}. For a one-hot encoded representation of $m$ ICD9 codes across our cohort, $\textbf{P} \in \mathbb{R}^{6470\times m}$, and an embedding matrix, $\textbf{E}\in \mathbb{R}^{m\times 300}$, our design matrix for clustering, $\textbf{X} \in \mathbb{R}^{6470\times 300}$,  is given by the following matrix multiplication:

$$\textbf{X} = \textbf{P} \cdot \textbf{E}.$$

This matrix multiplication sums the ICD embeddings across a patient record, and the resultant embedding is directly affected by the number of times each code appears in a patient's history. Another representation for the ICD9 codes was investigated, taking the row-wise mean of embeddings instead of their sum. This amounts to normalising $X$ by dividing by the row-wise sum of $\textbf{P}$. While these were compared, a more salient and interpretable clustering was observed for the sum of the embeddings for the ICD dataset. A comparison of silhouette scores and inertia also used in determining methodology and number of clusters can be seen in Supplementary Figure 1.

\subsubsection{Patient Notes}

\begin{figure}
    \centering
    \includegraphics[width=\textwidth]{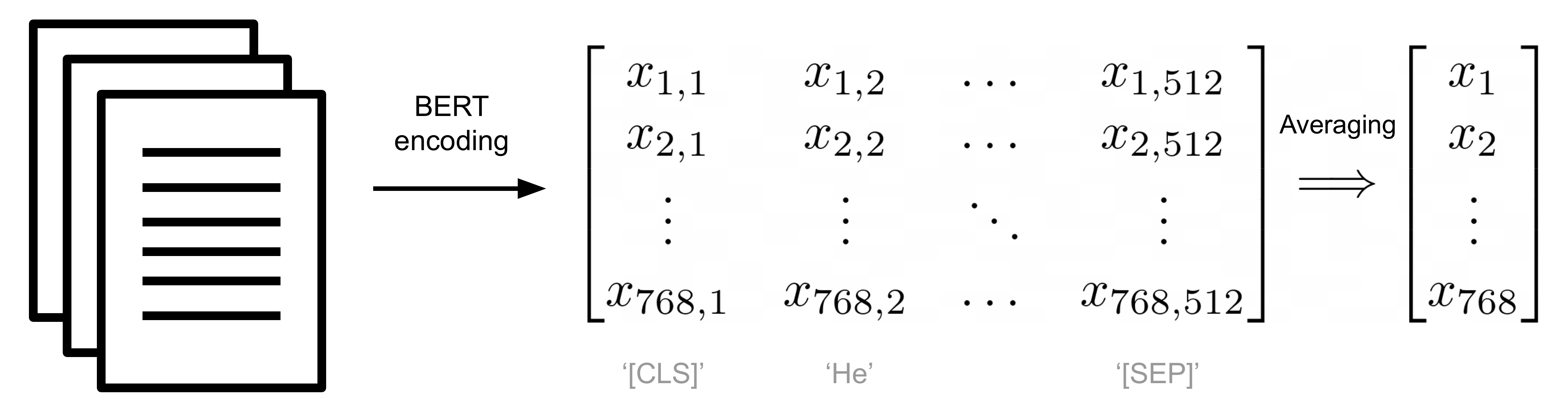}
    \caption{\textbf{Notes preprocessing pipeline vizualization.} A schematic diagram showing the note processing pipeline for a 512-token input sequence to Clinical BERT.}
    \label{fig:notes_processing}
\end{figure}

Patient notes were encoded using Clinical BERT prior to clustering. In order to derive patient-level representations from notes, a number of preprocessing steps were undertaken. First, unwanted delimiter characters were stripped from patient notes, and notes were chunked into contiguous sections of 1024 characters. This resulted in a distribution of token numbers of $\sim$200-300 per note following BERT's WordPiece encoding of input sequences. After passing through the transformer encoding, we took the final layer representation averaged over the 12 attention heads, such that each note was represented by a matrix of dimension $(768, n)$, corresponding to each of the $n$ input tokens having a 768-dimensional contextual vector representing it. Following this encoding, the representation was averaged over the token dimension, so as to arrive at a single 768-dimensional vector for the whole note. This was explored using both simple averaging over the token dimension, as well as attention-weighted averaging based on row-wise entropy of the final layer attention matrix, and attention-weighted averaging was used in patient representations due to the resultant lower inertia and increased silhouette score on average, and figures demonstrating this can be found in Supplementary Figure 2. A schematic depicting the note representation pipeline can be seen in Fig. \ref{fig:notes_processing}.

\subsubsection{Attention-weighted averaging}

For a given input sequence of length $n$, the final layer representation in a transformer model has an associated self-attention matrix, $\textbf{A} \in \mathbb{R}^{n \times n}$. For BERT-based models, $n \leq 512$ due to the constraint on the length of the input sequence. In order to provide weights for averaging the embedding, $\textbf{E} \in \mathbb{R}^{768 \times n}$, over the token dimension, we compute the row-wise differential entropy of the attention matrix, given in equation \ref{eq:diff_entropy}. Differential entropy is the continuous analogue of Shannon entropy, which is usually only defined for discrete random variables \cite{shannon1948mathematical}. In particular, the method described in Ebrahimi \textit{et al.} \cite{ebrahimi1994two} is used to approximate the differential entropy, implemented in the Python library SciPy  \cite{2020SciPy-NMeth}, as the closed-form expression for the attention distribution for a given row $f(x)$ is not known analytically from the values of attention sampled. The differential entropy for a row $i$ is given by

\begin{equation}\label{eq:diff_entropy}
    h_i = h(A_i) = - \int_{\mathcal{A}_i} f(x)log(f(x)) dx,
\end{equation}

\noindent
and the corresponding vector $\textbf{h} \in \mathbb{R}^{n \times 1}$ corresponds to the entropy across every row. From the row-wise entropy, this vector is softmaxed to get the corresponding weights, $\textbf{w} \in \mathbb{R}^{n\times 1}$:

\begin{equation}
    w_i = \sigma(\textbf{h})_i = \frac{e^{h_i}}{\sum\limits_{j=1}^{n}{e^{h_{j}}}}.
\end{equation}

\noindent
The resultant note embedding, \textbf{N}, is then given by the matrix multiplication

\begin{equation}
   \textbf{N} = \textbf{E} \cdot \textbf{w},
\end{equation}
 \noindent
and the final patient-level representation is the simple average over all note fragments for a given patient, over all of their encounters. A visualization of attention matrices with varying row-wise entropy and thus varying weighting is given in Supplementary Figures 3 and 4.

\subsection{Hierarchical Clustering}

Following the encoding of both ICD9 codes for the ICD data and clinical text for the notes data, a clustering analysis was performed for each representation. We employed hierarchical agglomerative clustering using Ward's linkage criterion, implemented in the Python package Scikit-learn \cite{scikit-learn}. Elbow plots and silhouette scores were used to compare clustering methods and to select the number of clusters for each of the two datasets. Following clustering, resultant clusters were characterised by testing for enrichment of ICD9 codes within each cluster. This enrichment analysis was performed for both the ICD representation and the patient notes representation, even though the codes themselves were not used in clustering the notes representation. Within each cluster, a 2x2 contingency table was generated for each ICD9 diagnosis code, comparing counts of patients with that code within-cluster to counts of patients with that same code in other clusters. The relative prevalence of each code as well as a fold-change between groups was characterised, and a $\chi^{2}$ test for enrichment was performed. A Bonferroni correction was applied to the resultant $p$-values to correct for multiple comparisons, and ICD codes with positive fold-change were ranked by corrected $p$-value to characterise the most significant enrichments. All $p$-values in this investigation were two-sided, with a post-corrected $\alpha$ of 0.05 to determine significance. For clusters derived in the notes clustering, enrichment of words within notes was also characterised by TF-IDF (term frequency–inverse document frequency). In this context, the sum total of all notes within a cluster is considered a document, and the set of stop words utilized was adapted from scikit-learn's English stop word list to include nonspecific medical terminology.

\section{Results}

\begin{table}[b!]
    \centering
    \begin{tabular}{ll}
    \toprule
        \textbf{Variable} & \textbf{Value}   \\
        ~ & (N = 6,470) \\ \midrule
        Age at first encounter & 72.6 ± 9.8  \\ 
        Age at last encounter & 75.2 ± 9.7  \\ 
        Female & 3,316 (51.3\%) \\ 
        Race (non-White) & 861 (13.3\%) \\ 
        Ethnicity (Hispanic) & 275 (4.3\%) \\ 
        With AD code & 2,970 (45.9\%)  \\ 
        With FTD code & 463 (7.2\%)  \\ 
        With LBD code & 359 (5.5\%)  \\ 
        With Vascular dementia code & 649 (10.0\%)  \\ 
        With MCI/Memory loss code & 3,922 (60.6\%) \\ 
        Number of unique codes & 36.1 ± 32.5 \\ 
        Number of patients genotyped & 487 (7.5\%)  \\ 
        APOE dosage & 0.55 ± 0.65\\ 
        \bottomrule \\
    \end{tabular}
    \caption{Patient cohort characteristics. Key: AD = Alzheimer's disease, FTD = Frontotemporal dementia, LBD = Lewy body dementias, MCI = Mild cognitive impairment.}
    \label{table:cohort_characteristics}
\end{table}

Table \ref{table:cohort_characteristics} shows the distribution of characteristics across our selected patient cohort. For this cohort, both ICD-based and notes-based representations from patient EHRs were derived and clustered independently. One thing to note is the prevalence of AD, as determined by having the relevant ICD code, is relatively low overall. While a significant number of patients lack an ICD diagnosis code for AD, this does not necessarily mean the prevalence of AD is low. Diagnosis is only probable until postmortem examination of brain tissue, and neurologists are typically reluctant to diagnose the condition until patients are further along in the disease course. Therefore, the substantial majority of patients seen in this cohort are likely to have AD or go on to be diagnosed with it at some point. A subset of patients are affected with other dementias, including frontotemporal dementia, Lewy body dementia, and vascular dementia. While the prevalence of vascular dementia is the highest among these, there is some overlap in diagnoses and many of these patients have an AD ICD code alongside their other dementia diagnosis.

\subsection{ICD Clustering}

\begin{figure}

\centering
\includegraphics[width=\textwidth]{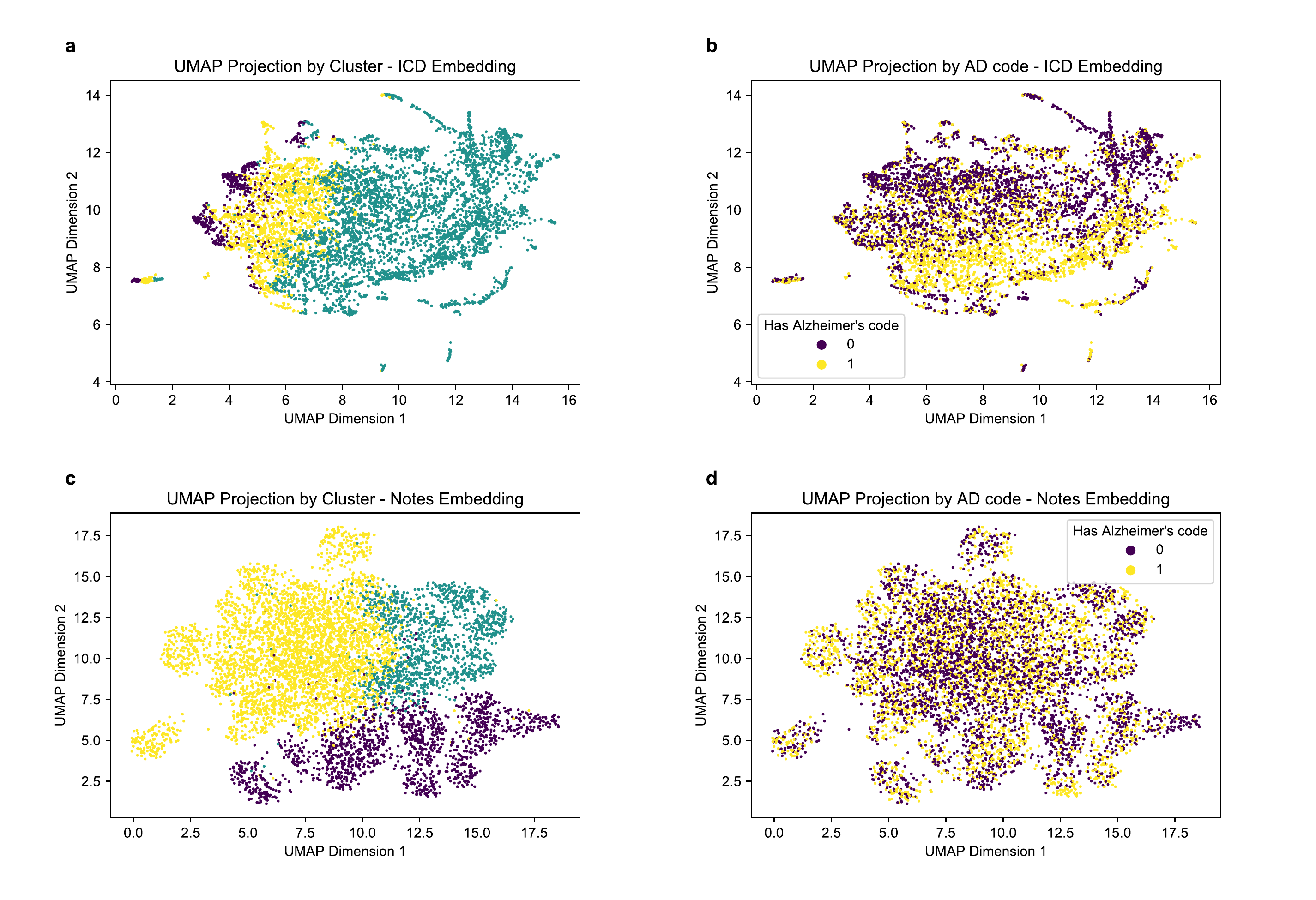}

\caption{\textbf{UMAP projections of patient cohort across both representations.} UMAP projections of the ICD representation coloured by cluster membership (a) and patients having Alzheimer's ICD9 code (b), as well as the corresponding projection for the patient notes representation (c, d).}
\label{fig:umap_clusters}
\end{figure}

Three clusters were identified in the ICD representation of patient EHRs. Fig. \ref{fig:umap_clusters}a shows a UMAP \cite{mcinnes2018umap} projection of the ICD code embeddings, colored by cluster. Fig. \ref{fig:umap_clusters}b shows the same projection but colored by the presence of an AD code, where it can be seen that the main axis of variation in clustering is approximately orthogonal to the axis of variation in AD code presence. This indicates that the phenotypic distinction captured in clustering is not entirely confounded by the presence of the AD code alone, and that clustering is therefore driven by other phenotypic features.

The first cluster ($N=508$) was characterised by ICD codes for chronic pain and type 2 diabetes (T2D). There was a 3-6 fold increase in prevalence of pain-related codes like foot pain and fibromyalgia compared to patients in other clusters, and approximately 20\% of patients within this cluster had T2D codes, compared with a background prevalence of ~3\% in the other clusters. This cluster was marginally more female and significantly more non-White than the other clusters. The prevalence of AD within this cluster was also lower, though with substantially higher prevalence of mild cognitive impairment or memory loss compared to the cohort average. This may indicate a group of patients earlier on in their timeline of disease progression and diagnosis, though with a larger burden of primary care as indicated by having an increased number of unique ICD codes compared to the cohort average. A higher proportion of patients were genotyped in this cluster but with a lower APOE dosage on average compared to other clusters.

A second cluster ($N=4,350$) was defined by lack of significant enrichment for the codes as found in the other clusters, and a marginally higher prevalence of neurodegenerative/focal neurological conditions, without being statistically significant, with the exception of frontotemporal dementia and Alzheimer's disease. This group of patients had lower numbers of unique codes, indicating less of a substantial interaction with the health system outside of their neurology visits.  This group also had a significantly lower prevalence of Hispanic patients.

\begin{figure}
    \centering
    \includegraphics[width=0.9\textwidth]{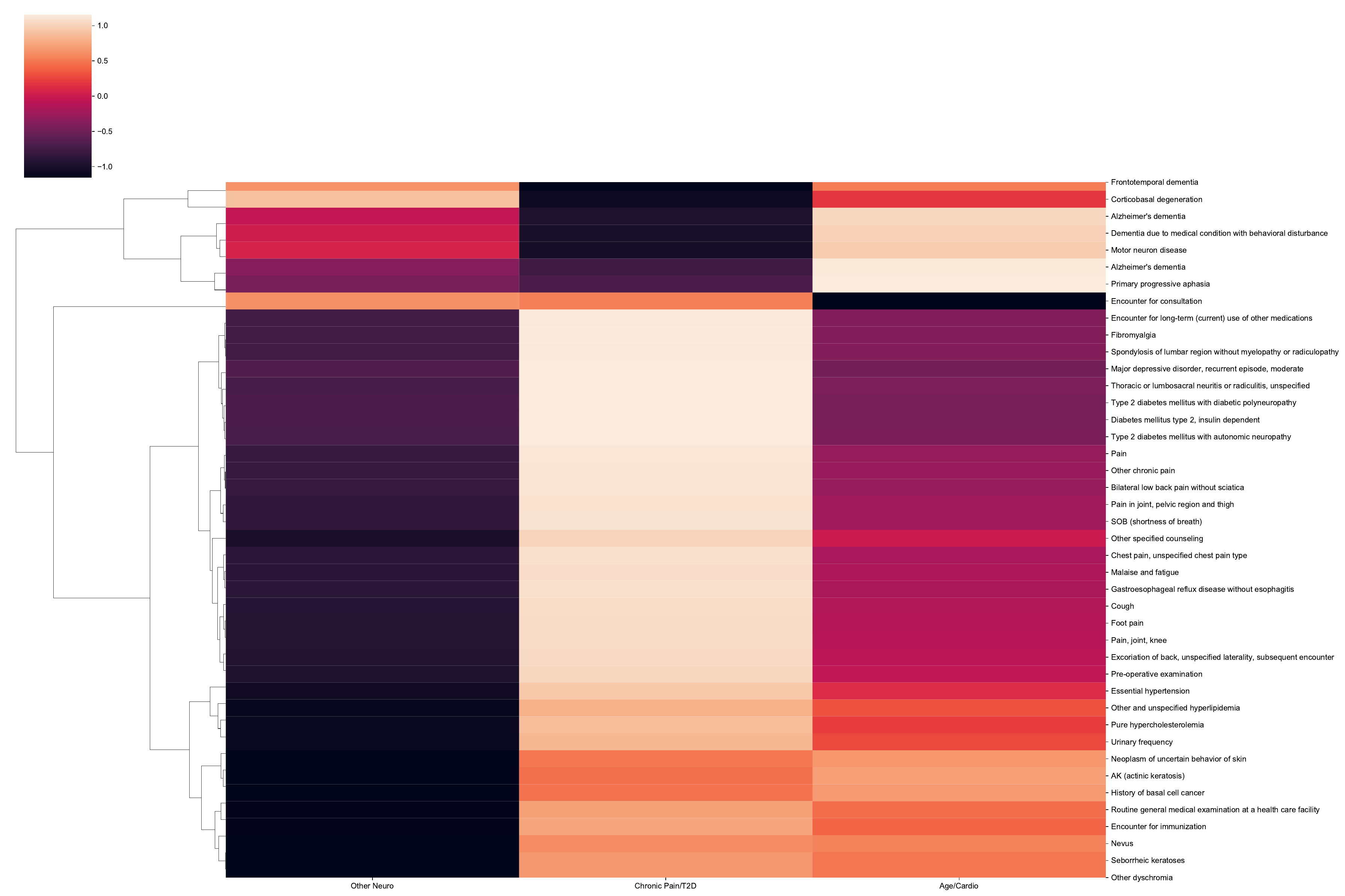}
    \caption{\textbf{ICD clustering heatmap.} A heatmap showing ICD9 code enrichment for clusters found in the ICD clustering. The colour of each code-cluster pair is determined by the normalized Z-score for that code.}
    \label{fig:icd_cluster_heatmap}
\end{figure}

The third and final cluster ($N=1,612$) was characterised by codes related to cardiovascular and general diseases of aging, as well as some overlap with the pain cohort. Patients within this cluster also had a more general interaction with the healthcare system overall, with enriched codes for routine general examinations and immunization. This may indicate a subset of patients that interacted with the MGB health system substantially for general primary care outside of the neurology department, which is consistent with the increased number of unique ICD codes compared to the cohort average. Patients in this cluster were significantly older at their first and last encounters, and had higher prevalence of vascular dementia and MCI/memory loss compared to the cohort average. Fig. \ref{fig:icd_cluster_heatmap} shows representative ICD codes for each of the three clusters, colored by their relative enrichment (row-wise $Z$-scores). Table. \ref{table:icd_cluster_table} provides a description of cluster characteristics for the ICD clustering, and tables depicting more detail on the specific ICD codes enriched within each cluster can be found in Supplementary Tables 1-6.

\begin{table}
    \centering
    \begin{tabular}{lllll}
    \toprule
        \textbf{Variable} & \textbf{Chronic Pain/T2D} & \textbf{Other Neuro} & \textbf{Cardio/Aging} & $p$-\textbf{value} \\ 
        ~ & (N = 508) & (N = 4,350) & (N = 1,612) &  \\ \midrule
        Age at first encounter & 72.4 ± 9.7 & 71.9 ± 9.9 & 74.7 ± 9.3 & \textbf{< 0.001}\textsuperscript{a} \\ 
        Age at last encounter & 75.0 ± 9.6 & 74.4 ± 9.8 & 77.3 ± 9.2 & \textbf{< 0.001}\textsuperscript{a} \\ 
        Female & 295 (58.1\%) & 2,236 (51.4\%) & 785 (48.7\%)    &  0.12\textsuperscript{b} \\ 
        Race (non-White) & 92 (18.1\%) & 544 (12.5\%) & 241 (15.0\%) & \textbf{0.003}\textsuperscript{b}  \\ 
        Ethnicity (Hispanic) & 31 (6.1\%) & 149 (3.4\%) & 95 (5.9\%) & \textbf{< 0.001}\textsuperscript{b} \\ 
        With AD code & 122 (24.0\%) & 2,177 (50.0\%) & 671 (41.6\%) & \textbf{< 0.001}\textsuperscript{b} \\ 
        With FTD code & 17 (3.3\%) & 372 (8.6\%) & 74 (4.6\%) & \textbf{< 0.001}\textsuperscript{b} \\ 
        With LBD code & 24 (4.7\%) & 243 (5.6\%) & 92 (5.7\%) & 0.714\textsuperscript{b} \\ 
        With Vascular dementia code & 64 (12.6\%) & 370 (8.5\%) & 215 (13.3\%)  & \textbf{< 0.001}\textsuperscript{b} \\ 
        With MCI/Memory loss code & 395 (77.8\%) & 2,303 (52.9\%) & 1,224 (75.9\%) & \textbf{< 0.001}\textsuperscript{b} \\ 
        Number of unique codes & 98.5 ± 36.3 & 19.9 ± 15.7 & 64.0 ± 21.5 & \textbf{< 0.001}\textsuperscript{a} \\ 
        Number of patients genotyped & 86 (16.9\%) & 228 (5.2\%) & 173 (10.7\%) & \textbf{< 0.001}\textsuperscript{b} \\ 
        APOE dosage & 0.40 ± 0.54 & 0.60 ± 0.69 & 0.57 ± 0.65 & \textbf{0.0045}\textsuperscript{a} \\ \bottomrule  \\
    \end{tabular}
    \caption{ICD clustering characteristics. Key: T2D = Type 2 diabetes, KD = Kidney disease, AD = Alzheimer's disease, FTD = Frontotemporal dementia, LBD = Lewy body dementias, MCI = Mild cognitive impairment. Statistical tests: $\textsuperscript{a}$ one-way ANOVA ($F$-test), $\textsuperscript{b}$ $\chi^2$ test of independence.}
    \label{table:icd_cluster_table}
\end{table}

\subsection{Note Cohort Clusters}

Three clusters were also identified in the patient notes clustering. As in the ICD clustering, Fig. \ref{fig:umap_clusters}c shows a UMAP projection of the note embeddings, colored by cluster, and Fig. \ref{fig:umap_clusters}d shows the same projection but colored by the presence of an AD code. Here we again see a clear boundary between clusters in the 2D projection, which appears relatively independent to the prevalence of AD as measured by ICD code. This indicates that the phenotypic distinction captured in clustering is not entirely confounded by the presence of the AD code alone, and that clustering is therefore driven by other phenotypic features.

The first cluster ($N=1,469$) was characterised by the enrichment for codes related to hydrocephalus, together with codes for Lewy body disease (both Lewy body dementias and Parkinson's disease). This indicates a patient cluster that overlaps phenotypically in terms of movement disorders and gait. It should be noted that while statistically significant with an approximate 5-fold enrichment, the prevalence of hydrocephalus and (as well as Parkinson's disease) within this cluster was still relatively low, at around 12\%.

The second cluster ($N=1,297$) identified in the notes clustering was characterised by significantly enriched codes for senile dementia and hypertension, and broadly overlapped in terms of other over-represented codes with the cardiovascular/aging cluster from the ICD clustering.  While not significant after correction for multiple comparisons, many other codes for cardiovascular and dermatological conditions exhibit increased prevalence, highlighting that this cluster was also distinguished by a substantial volume of interaction with the MGB system for primary care in addition to being seen by memory specialists. AD prevalence is higher in this cluster, together with a slightly increased average age at first and last encounter compared to the cohort average.

\begin{figure}
    \centering
    \includegraphics[width=1\textwidth]{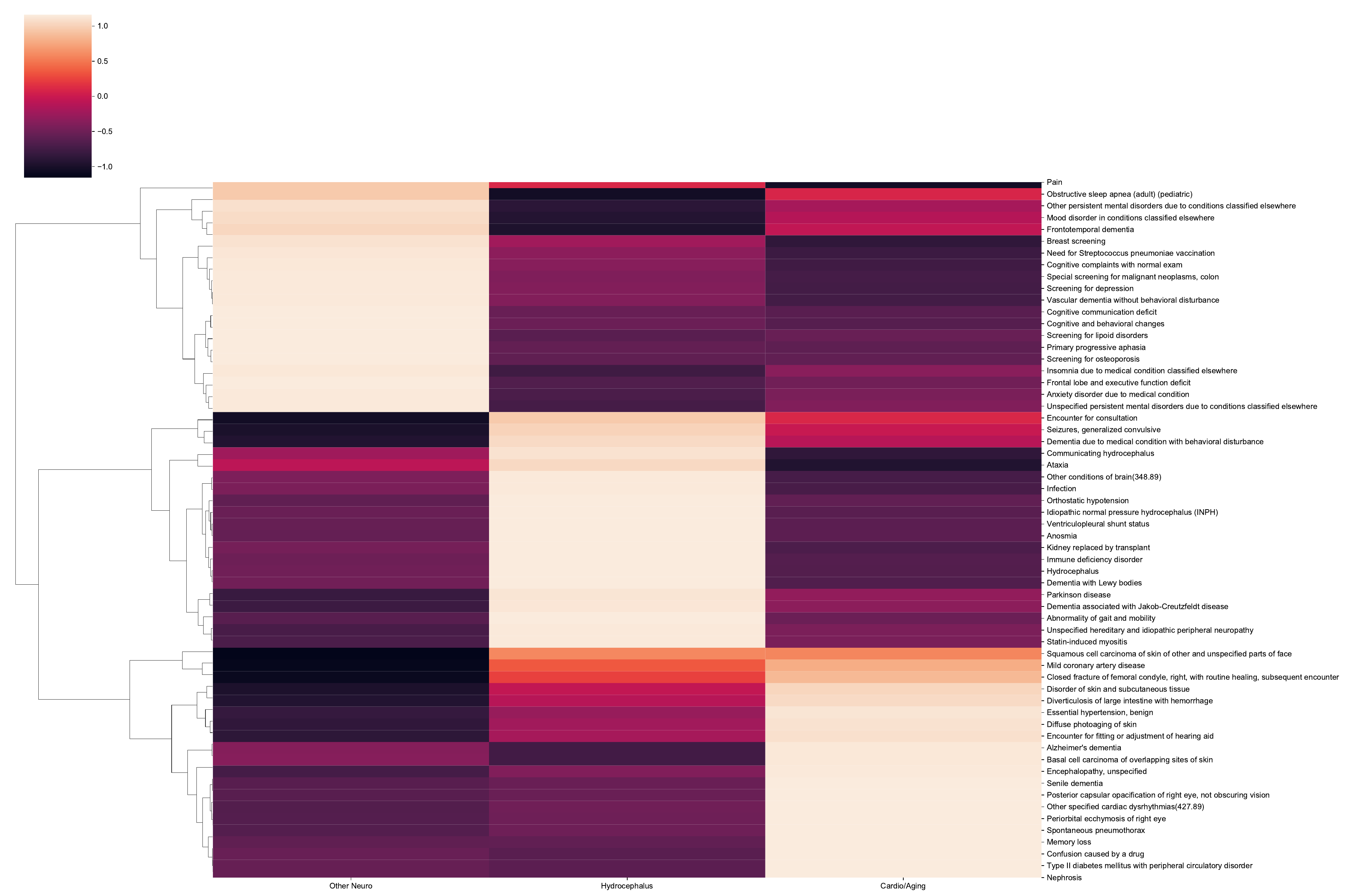}
    \caption{\textbf{Notes clustering heatmap.} A heatmap showing ICD9 code enrichment for clusters found in the patient notes clustering. The colour of each code-cluster pair is determined by the normalized Z-score for that code.}
    \label{fig:notes_heatmap}
\end{figure}

The third and final cluster ($N=3,704$) was characterised by a marginally higher prevalence of other neurodegenerative/focal neurological conditions, as also captured in clustering using the ICD representation. This cluster was also significantly enriched for neuropsychiatric conditions, including, mood disorders, anxiety disorder due to a medical condition, and screening for depression. Frontotemporal dementia was again significant, as well as other focal neurological conditions such as vascular dementia and primary progressive aphasia. Also of note for this cluster was the increased prevalence of non-White and Hispanic patients, as well as female patients. Representative codes for these three clusters are visualized in Fig. \ref{fig:notes_heatmap}, with this heatmap demonstrating a more disjoint separation in enriched codes than for the ICD clustering, and cluster characteristics for the notes clusters are shown in Table. \ref{table:notes_cluster_table}.

\begin{table}
    \centering
    \begin{tabular}{lllll}
    \toprule
        \textbf{Variable} & \textbf{Hydrocephalus/LBD} &\textbf{Cardio/Aging} & \textbf{Other Neuro} & \textbf{$p$-value}  \\
        ~ & (N = 1,469) & (N = 1,297) & (N = 3,704) & \\ \midrule
        Age at first encounter & 73.9 ± 9.5 & 74.7 ± 9.4 & 71.4 ± 9.9 & \textbf{< 0.001}\textsuperscript{a} \\ 
        Age at last encounter & 76.8 ± 9.1 & 77.5 ± 9.2 & 73.7 ± 9.9 & \textbf{< 0.001}\textsuperscript{a} \\ 
        Female & 696 (47.4\%) & 611 (47.1\%) & 2009 (54.2\%) & \textbf{0.007}\textsuperscript{b} \\  
        Race (non-White) & 147 (10.0\%) & 169 (13.0\%) & 561 (15.1\%)  & \textbf{< 0.001}\textsuperscript{b} \\ 
        Ethnicity (Hispanic) & 29 (2.0\%) & 48 (3.7\%) & 198 (5.3\%) & \textbf{< 0.001}\textsuperscript{b} \\ 
        With AD code & 638 (43.4\%) & 662 (51.0\%) & 1,670 (45.1\%) & \textbf{0.036}\textsuperscript{b} \\ 
        With FTD code & 57 (3.9\%) & 66 (5.1\%) & 340 (9.2\%) & \textbf{< 0.001}\textsuperscript{b} \\ 
        With LBD code & 155 (9.9\%) & 43 (3.3\%) & 161 (4.3\%) & \textbf{< 0.001}\textsuperscript{b} \\ 
        With Vascular dementia code & 133 (9.1\%) & 75 (5.8\%) & 441 (11.9\%) & \textbf{< 0.001}\textsuperscript{b}\\ 
        With MCI/Memory loss code & 860 (58.5\%) & 794 (61.2\%) & 2,268 (61.2\%) & 0.654\textsuperscript{b} \\ 
        Number of unique encounters & 4.9 ± 3.8 & 4.5 ± 3.9 & 4.9 ± 6.2 & 0.090\textsuperscript{a} \\ 
        Number of patients genotyped & 119 (8.1\%) & 114 (8.8\%) & 254 (6.9\%) & 0.075\textsuperscript{b} \\ 
        APOE dosage & 0.55 ± 0.65 & 0.51 ± 0.60 & 0.57 ± 0.68 & 0.664\textsuperscript{a} \\ \bottomrule \\
        
    \end{tabular}
    \caption{Patient notes clustering characteristics. Key: AD = Alzheimer's disease, FTD = Frontotemporal dementia, LBD = Lewy body dementias, MCI = Mild cognitive impairment. Statistical tests: $\textsuperscript{a}$ one-way ANOVA ($F$-test), $\textsuperscript{b}$ $\chi^2$ test of independence.}
    \label{table:notes_cluster_table}
\end{table}

In addition to the enrichment for ICD codes, the notes-based clustering was characterised by over-representation for individual words as measured by the TF-IDF statistic, as shown in Table. \ref{table:tf-idftable}. Despite some overlap between clusters, important words identified in each cluster by TF-IDF were generally concordant with their characterisation from the ICD code enrichment. Within the hydrocephalus/LBD cluster, important words included `pain' and `tremor`, which are both specific features of that phenotype. The cardiovascular/aging cluster identified words like `scan', `memantine', and `donepezil', which indicates diagnostic scanning as well as administration of common dementia medications. For the `other neuro' cluster, `language', `speech', and `word' are important, with speech-related symptoms being prevalent in many neurodegenerative conditions, as well as `anxiety' and `behavioral' corresponding to the neuropsychiatric conditions identified as prevalent in this cluster.

\begin{table}
    \centering
    \begin{tabular}{lll}
    \toprule
        \textbf{Hydrocephalus/LBD} & \textbf{Cardio/Aging} & \textbf{Other Neuro} \\ \midrule
        
        bp	& recall & 	social \\
        vitamin	& scan & anxiety \\
        tremor & 	decline & 	speech \\
        oral & donepezil & 	language \\
        nightly & memantine & behavioral \\
        route & behavioral & bilaterally \\
        active & work & pain \\
        outpatient & driving & word \\
        pain & social & recall \\
        mood & language & attention \\\bottomrule \\
        
    \end{tabular}
    \caption{TF-IDF for enriched words within each cluster.}
    \label{table:tf-idftable}
\end{table}

\section{Discussion}

The aim of this study was to characterise heterogeneity in Alzheimer's disease and related dementias, by applying existing representation learning techniques to patient EHRs. To that end, we have leveraged pre-trained embeddings for ICD codes and a state-of-the-art transformer architecture to encode patient notes, which have not been applied before in the context of unsupervised learning on EHRs for AD. In doing so, we have identified a number of well-defined phenotypes based on prevalent ICD codes within the resultant clusters.

These clusters arose after deriving two independent representations on the basis of data modality, one from structured patient ICD codes, and one from unstructured neurology notes. We identified three clusters in both the ICD representation and the patient notes representation, with a substantial degree of overlap in identified clusters between the two modalities. In the ICD clustering, a cluster was identified that was clearly characterised by the presence of codes related to chronic pain and type 2 diabetes. This cluster maps to a T2D cluster identified in Xu \textit{et al.} \cite{xu2020data} and the association between T2D and AD is well-known. T2D appears to be a risk factor in both AD onset and a faster rate of progression, hypothesized to be due to common causal pathways between the two diseases  \cite{arvanitakis2004diabetes, ravona2010diabetes, sims2010does}. While chronic pain does not emerge clearly in prior works clustering AD on EHRs, there does appear to be a relationship between chronic pain and AD/dementia \cite{whitlock2017association, cao2019link}. Any hypothetical causal link has yet to be established in this relationship, though it does appear that there is a bi-directional correlation. If chronic pain does indeed accelerate AD pathogenesis, then this would certainly warrant further investigation as a distinct phenotype of AD. 


Cardiovascular or cerebrovascular disease clusters were found to some extent in all prior works clustering AD patient EHRs, and in combination with general diseases of aging and dermatological comorbidities, were enriched in clusters identified in this work, for both data modalities. For example, hypertension, hyperlipidemia, and dysrhythmias were reliably identified in one or both clustering analyses. With a number of common risk factors, there is a well-established link between cardiovascular disease and AD, as well as in other dementias \cite{martins2006apolipoprotein, snyder2015vascular}. In addition to cardiovascular disease, an increased prevalence of dermatological conditions were identified in this cluster in both analyses. It is unclear whether there is a general relationship between dermatological conditions and AD, though one review has explored the potential relationship  \cite{zhang2021relationship}. Seborrheic keratosis was enriched for this cluster in the ICD clustering, and one study has linked its onset to overexpression of amyloid precursor protein, which plays an important role in AD pathogenesis  \cite{li2018overexpression}. For the ICD clustering, some chronic pain codes were also enriched in this cluster, though co-morbid with other diseases of aging, indicating some phenotypic overlap.

In the notes clustering, a cluster was found that was well-characterised by the prevalence of hydrocephalus as well as Lewy body disease (Lewy body dementia and Parkinson's). Hydrocephalus is a condition that results from a build-up of cerebrospinal fluid in the ventricles of the brain, and usually responds well to the placement of a shunt to drain excess fluid. While this cluster may not be representative of a distinct phenotype of AD, prior work has highlighted AD as a common source of comorbidity in patients with normal pressure hydrocephalus, and that AD status may limit the efficacy of shunt placement in patients presenting with both conditions \cite{hamilton2010lack, cabral2011frequency}. It has been suggested that the glymphatic (glia-lymphatic) system could be implicated the relationship between AD and hydrocephalus  \cite{reeves2020glymphatic}. The identification of this cluster may also represent a common diagnostic workflow, whereby the initial presentation for Lewy body disease may mimic that of hydrocephalus, with the conditions being common differential diagnoses for one another. 

This study is limited by the lack of an independent validation cohort in which to validate the identified clusters. While this limitation is somewhat mitigated by the relative overlap in identified clusters between the ICD and notes representations, as well as the fact that many of the clusters identified here were also found in prior work, applying the same encoding and clustering technique on a hold-out validation cohort would strengthen the results presented here.

Furthermore, this study is limited by the lack of clear signal regarding which patients should belong to the AD phenotype, which is a by-product of the billing-motivated nature of coding, as well as the underlying diagnostic complexity associated with this disease and its related dementias. Instead of relying on the presence of an AD ICD code in a patient record to define the phenotype as done in previous work, which has its own limitations, we instead have used the cohort selection process to inform the definition of our target phenotype. While the majority of patients identified will be probable AD patients, this required expanding the focus of disease heterogeneity from just AD to related dementias as well.

Finally, this work could be improved by jointly leveraging the two data modalities utilised here, as well as expanding to more structured and unstructured data like medications and imaging. With a greater availability of multi-modal data, methodological improvements could be made by training a deep autoencoder to jointly leverage modalities, rather than performing two clustering analyses in parallel, and relying on the heuristic averaging employed here. In this framework, explicitly leveraging temporal or graph properties of EHRs may contribute to more informative representations and therefore improved unsupervised clustering, as prior approaches have demonstrated in a supervised learning setting  \cite{zhu2021variationally, choi2020learning}.

\newpage

\bibliographystyle{unsrt}  
\bibliography{references}  

\end{document}


\section*{Supplementary Material}\label{sec:sup}

\subsection*{Enriched ICD9 codes in ICD Clusters}

The following tables show the top-20 ICD codes for each of the clusters identified in both the ICD and patient notes clustering analyses. They also provide information on the significance and relative prevalence of each of the codes within and between clusters.

\noindent\textbf{Key:}

\begin{itemize}
    \item \textbf{ICD Code}: ICD9 code
    \item \textbf{p\_value}: $p$-value from $\chi^2$ test. 
    \item \textbf{p\_value\_fdr}: Bonferroni-corrected $p$-value.
    \item \textbf{Sig.}: Statistical significance following correction for multiple comparisons, \\ *** $<$ 0.001; ** $<0.01$; * $<0.05$.
    \item \textbf{Prev.}: Prevalence of code within cluster.
    \item \textbf{Exp. Prev.}: Expected prevalence, prevalence of code in other clusters.
    \item \textbf{fc}: Fold-change between observed prevalence and expected prevalence.
    \item \textbf{DiagnosisNM}: ICD9 code name.
\end{itemize}

\begin{landscape}

\begin{table}[H]
    \centering
    \begin{tabular}{llllllll}
    \toprule
        \textbf{ICD Code} &\textbf{p\_value} & \textbf{p\_value\_fdr} & \textbf{Sig.} & \textbf{Prev.} & \textbf{Exp. Prev.} & \textbf{fc} & \textbf{DiagnosisNM}  \\ \midrule
        338.29 & 4.80E-122 & 2.74E-118 & *** & 0.62598 & 0.17930 & 3.491 & Other chronic pain \\
        724.2 & 8.10E-116 & 4.63E-112 & *** & 0.54724 & 0.14391 & 3.803 & Bilateral low back pain without sciatica \\
        729.5 & 8.45E-110 & 4.83E-106 & *** & 0.61417 & 0.18668 & 3.290 & Foot pain \\
        729.1 & 2.99E-98 & 1.71E-94 & *** & 0.30118 & 0.05099 & 5.907 & Fibromyalgia \\
        721.3 & 1.45E-97 & 8.31E-94 & *** & 0.28346 & 0.04529 & 6.259 & Spondylosis of lumbar region without myelopathy \\
        & & & & & & & or radiculopathy \\ 
        786.2 & 6.60E-97 & 3.77E-93 & *** & 0.55118 & 0.16639 & 3.313 & Cough \\ 
        724.4 & 1.60E-96 & 9.15E-93 & *** & 0.29331 & 0.04914 & 5.968 & Thoracic or lumbosacral neuritis or radiculitis, unspecified \\ 
        780.79 & 1.86E-94 & 1.07E-90 & *** & 0.59055 & 0.19239 & 3.070 & Malaise and fatigue \\ 
        357.2 & 2.93E-93 & 1.68E-89 & *** & 0.21063 & 0.02499 & 8.428 & Type 2 diabetes mellitus with diabetic polyneuropathy \\ 
        719.45 & 3.31E-90 & 1.89E-86 & *** & 0.38583 & 0.08990 & 4.292 & Pain in joint, pelvic region and thigh \\ 
        V58.67 & 3.91E-89 & 2.24E-85 & *** & 0.21457 & 0.02751 & 7.800 & Diabetes mellitus type 2, insulin dependent \\ 
        V58.69 & 4.12E-87 & 2.36E-83 & *** & 0.32480 & 0.06642 & 4.890 & Encounter for long-term (current) use of other medications \\ 
        530.81 & 3.20E-84 & 1.83E-80 & *** & 0.44291 & 0.12278 & 3.607 & Gastroesophageal reflux disease without esophagitis \\ 
        786.05 & 1.70E-80 & 9.69E-77 & *** & 0.32480 & 0.07112 & 4.567 & SOB (shortness of breath) \\ 
        V58.89 & 2.04E-79 & 1.16E-75 & *** & 0.43504 & 0.12395 & 3.510 & Excoriation of back, unspecified laterality,\\
        & & & & & & & subsequent encounter \\ 
        250.60 & 2.99E-79 & 1.71E-75 & *** & 0.21063 & 0.03019 & 6.977 & Type 2 diabetes mellitus with autonomic neuropathy \\ 
        780.96 & 1.70E-78 & 9.69E-75 & *** & 0.38583 & 0.10030 & 3.847 & Pain \\ 
        296.32 & 5.01E-77 & 2.87E-73 & *** & 0.20866 & 0.03053 & 6.835 & Major depressive disorder, recurrent episode, moderate \\ 
        V65.49 & 1.24E-75 & 7.09E-72 & *** & 0.48425 & 0.15548 & 3.114 & Other specified counseling \\ 
        786.50 & 1.89E-74 & 1.08E-70 & *** & 0.36220 & 0.09292 & 3.898 & Chest pain, unspecified chest pain type \\ 
        \bottomrule \\
    \end{tabular}
    \caption{Top 20 enriched codes for chronic pain/T2D cluster in the ICD clustering.}
    \label{table:icd_chronic_pain}
\end{table}

\begin{table}[H]
    \centering
    \begin{tabular}{llllllll}
    \toprule
        \textbf{ICD Code} &\textbf{p\_value} & \textbf{p\_value\_fdr} & \textbf{Sig.} & \textbf{Prev.} & \textbf{Exp. Prev.} & \textbf{fc} & \textbf{DiagnosisNM}  \\ \midrule
331.0 & 1.30E-21 & 7.42E-18 & *** & 0.50046 & 0.37406 & 1.338 & Alzheimer's dementia \\ 
        294.10 & 1.86E-12 & 1.06E-08 & *** & 0.48621 & 0.39292 & 1.237 & Alzheimer's dementia \\ 
        331.19 & 1.90E-10 & 1.09E-06 & *** & 0.07678 & 0.03538 & 2.170 & Frontotemporal dementia \\
        294.11 & 8.34E-08 & 0.000476948 & *** & 0.25540 & 0.19481 & 1.311 & Dementia due to medical condition \\
        & & & & & & &  with behavioral disturbance \\ 
        V65.9 & 6.45E-07 & 0.003687954 & ** & 0.08943 & 0.05377 & 1.663 & Encounter for consultation \\ 
        335.20 & 0.006012318 & 34.37843242 & ~ & 0.01080 & 0.00377 & 2.863 & Motor neuron disease \\ 
        331.6 & 0.011095009 & 63.44126254 & ~ & 0.00943 & 0.00330 & 2.855 & Corticobasal degeneration \\ 
        784.3 & 0.020494271 & 117.1862438 & ~ & 0.07494 & 0.05896 & 1.271 & Primary progressive aphasia \\ 
        333.0 & 0.141054986 & 806.5524072 & ~ & 0.01747 & 0.01226 & 1.425 & Neurogenic orthostatic hypotension \\ 
        784.69 & 0.174142823 & 995.7486616 & ~ & 0.01977 & 0.01462 & 1.352 & Apraxia \\ 
        293.84 & 0.198053555 & 1132.470228 & ~ & 0.02322 & 0.01792 & 1.295 & Anxiety disorder due to medical condition \\ 
        290.20 & 0.206186693 & 1178.975511 & ~ & 0.00483 & 0.00236 & 2.047 & Dementia, senile with delusions \\ 
        345.91 & 0.230943349 & 1320.534071 & ~ & 0.00207 & 0.00047 & 4.386 & Intractable epilepsy without status epilepticus \\ 
        331.11 & 0.247924607 & 1417.632902 & ~ & 0.01310 & 0.00943 & 1.389 & Pick's disease \\ 
        334.2 & 0.277332277 & 1585.785958 & ~ & 0.00851 & 0.00566 & 1.503 & Corticobasal syndrome \\ 
        331.9 & 0.286177934 & 1636.365427 & ~ & 0.02851 & 0.02358 & 1.209 & Cerebral degeneration, unspecified \\ 
        046.19 & 0.352705118 & 2016.767865 & ~ & 0.00299 & 0.00142 & 2.112 & Dementia associated with Jakob-Creutzfeldt disease \\ 
        742.3 & 0.39801109 & 2275.827412 & ~ & 0.00161 & 0.00047 & 3.411 & Congenital hydrocephalus \\ 
        319 & 0.477699593 & 2731.486275 & ~ & 0.00207 & 0.00094 & 2.193 & Mental retardation \\ 
        783.4 & 0.522523203 & 2987.787672 & ~ & 0.00138 & 0.00047 & 2.924 & Development delay \\ \bottomrule \\
    \end{tabular}
    \caption{Top 20 enriched codes for other neuro cluster in ICD clustering.}
    \label{table:icd_other_neuro}
\end{table}

\begin{table}[H]
    \centering
    \begin{tabular}{lllllllll}
    \toprule
        \textbf{ICD Code} &\textbf{p\_value} & \textbf{p\_value\_fdr} & \textbf{Sig.} & \textbf{Prev.} & \textbf{Exp. Prev.} & \textbf{fc} & \textbf{DiagnosisNM}  \\ \midrule
        V70.0 & 6.52E-189 & 3.73E-185 & *** & 0.71774 & 0.30342 & 2.366 & Routine general medical examination  \\ 
        & & & & & & & at a health care facility \\ 
        401.9 & 2.44E-138 & 1.39E-134 & *** & 0.70906 & 0.35138 & 2.018 & Essential hypertension \\ 
        702.0 & 3.06E-130 & 1.75E-126 & *** & 0.33685 & 0.08748 & 3.850 & AK (actinic keratosis) \\ 
        702.19 & 1.56E-126 & 8.91E-123 & *** & 0.39826 & 0.12598 & 3.161 & Seborrheic keratoses \\ 
        338.29 & 1.44E-124 & 8.23E-121 & *** & 0.42494 & 0.14450 & 2.941 & Other chronic pain \\ 
        729.5 & 9.46E-120 & 5.41E-116 & *** & 0.42866 & 0.15109 & 2.837 & Foot pain \\ 
        272.4 & 1.79E-111 & 1.03E-107 & *** & 0.52481 & 0.22828 & 2.299 & Other and unspecified hyperlipidemia \\ 
        786.2 & 2.20E-99 & 1.26E-95 & *** & 0.37841 & 0.13627 & 2.777 & Cough \\
        238.2 & 5.25E-93 & 3.00E-89 & *** & 0.25993 & 0.07019 & 3.703 & Neoplasm of uncertain behavior of skin \\ 
        780.79 & 1.18E-92 & 6.75E-89 & *** & 0.40757 & 0.16262 & 2.506 & Malaise and fatigue \\ 
        V10.83 & 7.82E-85 & 4.47E-81 & *** & 0.23697 & 0.06340 & 3.738 & History of basal cell cancer \\ 
        709.09 & 3.66E-84 & 2.10E-80 & *** & 0.27295 & 0.08399 & 3.250 & Other dyschromia \\ 
        216.9 & 3.89E-83 & 2.22E-79 & *** & 0.24504 & 0.06896 & 3.553 & Nevus \\ 
        V03.89 & 1.89E-80 & 1.08E-76 & *** & 0.80397 & 0.53643 & 1.499 & Encounter for immunization \\ 
        719.46 & 1.05E-75 & 6.03E-72 & *** & 0.29467 & 0.10436 & 2.823 & Pain, joint, knee \\ 
        724.2 & 2.37E-74 & 1.36E-70 & *** & 0.32568 & 0.12577 & 2.589 & Bilateral low back pain without sciatica \\ 
        272.0 & 1.07E-73 & 6.11E-70 & *** & 0.37965 & 0.16385 & 2.317 & Pure hypercholesterolemia \\ 
        V72.84 & 1.92E-73 & 1.10E-69 & *** & 0.30769 & 0.11486 & 2.679 & Pre-operative examination \\ 
        V58.89 & 5.43E-71 & 3.10E-67 & *** & 0.28536 & 0.10292 & 2.773 & Excoriation of back, unspecified laterality, \\
        & & & & & & & subsequent encounter \\ 
        788.41 & 9.11E-70 & 5.21E-66 & *** & 0.25558 & 0.08563 & 2.985 & Urinary frequency \\ \bottomrule \\
    \end{tabular}
    \caption{Top 20 enriched codes for cardiovascular/aging cluster in ICD clustering.}
    \label{table:icd_cardio_derm}
\end{table}

\subsection*{Enriched ICD9 codes in Notes Clusters}

\begin{table}[H]
    \centering
    \begin{tabular}{llllllll}
    \toprule
        \textbf{ICD Code} &\textbf{p\_value} & \textbf{p\_value\_fdr} & \textbf{Sig.} & \textbf{Prev.} & \textbf{Exp. Prev.} & \textbf{fc} & \textbf{DiagnosisNM}  \\ \midrule
        331.5 & 8.44E-60 & 4.36E-56 & *** & 0.12594 & 0.02340 & 5.383 & Idiopathic normal pressure hydrocephalus (INPH) \\ 
        V45.2 & 9.08E-30 & 4.69E-26 & *** & 0.04629 & 0.00540 & 8.574 & Ventriculopleural shunt status \\ 
        331.4 & 5.83E-27 & 3.01E-23 & *** & 0.05786 & 0.01100 & 5.261 & Hydrocephalus \\ 
        332.0 & 1.26E-21 & 6.50E-18 & *** & 0.13342 & 0.05819 & 2.293 & Parkinson disease \\ 
        331.82 & 3.03E-21 & 1.56E-17 & *** & 0.10551 & 0.04079 & 2.587 & Dementia with Lewy bodies \\ 
        348.89 & 5.34E-20 & 2.76E-16 & *** & 0.05718 & 0.01480 & 3.864 & Other conditions of brain(348.89) \\ 
        781.2 & 7.18E-14 & 3.71E-10 & *** & 0.27093 & 0.18136 & 1.494 & Abnormality of gait and mobility \\ 
        331.3 & 2.10E-10 & 1.08E-06 & *** & 0.02519 & 0.00580 & 4.343 & Communicating hydrocephalus \\ 
        V65.9 & 4.72E-06 & 0.024394252 & * & 0.10619 & 0.06939 & 1.530 & Encounter for consultation \\ 
        781.3 & 0.000315501 & 1.629247132 & ~ & 0.04969 & 0.02979 & 1.668 & Ataxia \\ 
        046.19 & 0.000455391 & 2.351639622 & ~ & 0.00681 & 0.00120 & 5.674 & Dementia associated with Jakob-Creutzfeldt disease \\ 
        356.9 & 0.000559522 & 2.889370481 & ~ & 0.06263 & 0.04079 & 1.535 & Unspecified hereditary and idiopathic \\
        &  &  & & & & &  peripheral neuropathy \\ 
        458.0 & 0.001249156 & 6.450643386 & ~ & 0.04221 & 0.02559 & 1.649 & Orthostatic hypotension \\ 
        780.39 & 0.001310365 & 6.766726181 & ~ & 0.08237 & 0.05859 & 1.406 & Seizures, generalized convulsive \\ 
        136.9 & 0.001746166 & 9.01719964 & ~ & 0.00817 & 0.00220 & 3.714 & Infection \\ 
        E942.2 & 0.002219494 & 11.46146762 & ~ & 0.00340 & 0.00020 & 17.022 & Statin-induced myositis \\ 
        279.3 & 0.002219494 & 11.46146762 & ~ & 0.00340 & 0.00020 & 17.022 & Immune deficiency disorder \\ 
        781.1 & 0.003508 & 18.11531209 & ~ & 0.02110 & 0.01080 & 1.954 & Anosmia \\ 
        V42.0 & 0.003635117 & 18.77174336 & ~ & 0.00613 & 0.00140 & 4.377 & Kidney replaced by transplant \\ 
        294.11 & 0.003721492 & 19.21778421 & ~ & 0.26413 & 0.22715 & 1.163 & Dementia due to medical condition \\
        &  & & &  &  & &  with behavioral disturbance \\ \bottomrule \\
    \end{tabular}
    \caption{Top 20 enriched codes for hydrocephalus/LBD cluster in notes clustering.}
    \label{table:notes_lbd}
\end{table}

\begin{table}[H]
    \centering
    \begin{tabular}{llllllll}
    \toprule
       \textbf{ICD Code} &\textbf{p\_value} & \textbf{p\_value\_fdr} & \textbf{Sig.} & \textbf{Prev.} & \textbf{Exp. Prev.} & \textbf{fc} & \textbf{DiagnosisNM}  \\ \midrule
        290.0 & 2.29E-100 & 1.18E-96 & *** & 0.23978 & 0.05123 & 4.681 & Senile dementia \\ 
        401.1 & 2.62E-06 & 0.013535797 & * & 0.17656 & 0.12604 & 1.401 & Essential hypertension, benign \\ 
        331.0 & 3.78E-05 & 0.195187341 & ~ & 0.51041 & 0.44616 & 1.144 & Alzheimer's dementia \\ 
        692.74 & 4.95E-05 & 0.255842387 & ~ & 0.04318 & 0.02242 & 1.925 & Diffuse photoaging of skin \\ 
        348.30 & 0.000137693 & 0.711045922 & ~ & 0.03701 & 0.01894 & 1.954 & Encephalopathy, unspecified \\ 
        709.9 & 0.000431726 & 2.22943438 & ~ & 0.10100 & 0.07133 & 1.416 & Disorder of skin and subcutaneous tissue \\ 
        427.89 & 0.000502806 & 2.59649085 & ~ & 0.09792 & 0.06901 & 1.419 & Other specified cardiac dysrhythmias(427.89) \\ 
        562.12 & 0.000575282 & 2.970755233 & ~ & 0.00463 & 0.00039 & 11.965 & Diverticulosis of large intestine with hemorrhage \\ 
        414.00 & 0.000607857 & 3.138971122 & ~ & 0.12953 & 0.09666 & 1.340 & Mild coronary artery disease \\ 
        921.1 & 0.000936006 & 4.833534464 & ~ & 0.00694 & 0.00135 & 5.128 & Periorbital ecchymosis of right eye \\ 
        V53.2 & 0.001189183 & 6.140943366 & ~ & 0.02699 & 0.01373 & 1.966 & Encounter for fitting or adjustment of hearing aid \\ 
        443.81 & 0.001993231 & 10.29304402 & ~ & 0.02082 & 0.00986 & 2.112 & Type II diabetes mellitus with peripheral \\
        &  & &  & & & & circulatory disorder \\ 
        780.93 & 0.002661917 & 13.74613804 & ~ & 0.50810 & 0.46105 & 1.102 & Memory loss \\ 
        512.89 & 0.003442329 & 17.77618542 & ~ & 0.00386 & 0.00039 & 9.971 & Spontaneous pneumothorax \\ 
        366.52 & 0.00366121 & 18.90648649 & ~ & 0.00848 & 0.00251 & 3.375 & Posterior capsular opacification of right eye, \\ 
        &  & & &  &  & &  not obscuring vision  \\ 
        173.32 & 0.005218198 & 26.94677658 & ~ & 0.02082 & 0.01063 & 1.958 & Squamous cell carcinoma of skin of other \\ 
        &  & & &  &  & &  and unspecified parts of face  \\ 
        173.81 & 0.005252131 & 27.12200313 & ~ & 0.00308 & 0.00019 & 15.954 & Basal cell carcinoma of overlapping sites of skin \\ 
        292.81 & 0.005252131 & 27.12200313 & ~ & 0.00308 & 0.00019 & 15.954 & Confusion caused by a drug \\ 
        581.9 & 0.005252131 & 27.12200313 & ~ & 0.00308 & 0.00019 & 15.954 & Nephrosis \\
        V54.15 & 0.007277841 & 37.58277194 & ~ & 0.00617 & 0.00155 & 3.988 & Closed fracture of femoral condyle, right, \\ 
        &  & & &  &  & &  with routine healing, subsequent encounter \\ \bottomrule \\
    \end{tabular}
    \caption{Top 20 enriched codes for cardiovascular/aging cluster in notes clustering.}
    \label{table:notes_cardio_derm}
\end{table}

\begin{table}[H]
    \centering
    \begin{tabular}{llllllll}
    \toprule
        \textbf{ICD Code} &\textbf{p\_value} & \textbf{p\_value\_fdr} & \textbf{Sig.} & \textbf{Prev.} & \textbf{Exp. Prev.} & \textbf{fc} & \textbf{DiagnosisNM}  \\ \midrule
        799.59 & 2.97E-20 & 1.53E-16 & *** & 0.17846 & 0.09689 & 1.842 & Cognitive complaints with normal exam \\ 
        293.83 & 1.36E-19 & 7.02E-16 & *** & 0.05562 & 0.01229 & 4.524 & Mood disorder in conditions classified elsewhere \\
        293.84 & 2.27E-19 & 1.17E-15 & *** & 0.03564 & 0.00253 & 14.082 & Anxiety disorder due to medical condition \\ 
        294.9 & 3.88E-18 & 2.00E-14 & *** & 0.32046 & 0.22234 & 1.441 & Unspecified persistent mental disorders due to conditions \\ 
        & & & & & & & classified elsewhere \\ 
        784.3 & 3.81E-17 & 1.97E-13 & *** & 0.09287 & 0.03868 & 2.401 & Primary progressive aphasia \\ 
        V79.0 & 3.65E-15 & 1.89E-11 & *** & 0.07208 & 0.02748 & 2.623 & Screening for depression \\ 
        294.8 & 1.42E-11 & 7.33E-08 & *** & 0.10367 & 0.05640 & 1.838 & Other persistent mental disorders due to conditions\\ 
        & & & & & & & classified elsewhere \\ 
        331.19 & 6.06E-11 & 3.13E-07 & *** & 0.08045 & 0.04013 & 2.005 & Frontotemporal dementia \\ 
        799.52 & 8.60E-10 & 4.44E-06 & *** & 0.05130 & 0.02133 & 2.405 & Cognitive communication deficit \\ 
        290.40 & 6.89E-09 & 3.56E-05 & *** & 0.11663 & 0.07303 & 1.597 & Vascular dementia without behavioral disturbance \\ 
        V76.10 & 2.26E-08 & 0.000116505 & *** & 0.13661 & 0.09111 & 1.499 & Breast screening \\ 
        V77.91 & 2.54E-07 & 0.001311741 & ** & 0.06263 & 0.03398 & 1.843 & Screening for lipoid disorders \\ 
        312.9 & 1.21E-06 & 0.00623477 & ** & 0.05022 & 0.02603 & 1.929 & Cognitive and behavioral changes \\ 
        780.96 & 1.43E-06 & 0.007383663 & ** & 0.13985 & 0.09978 & 1.402 & Pain \\ 
        327.01 & 1.69E-06 & 0.0087282 & ** & 0.02781 & 0.01048 & 2.652 & Insomnia due to medical condition classified elsewhere \\ 
        V76.51 & 1.89E-06 & 0.009737959 & ** & 0.09125 & 0.05893 & 1.548 & Special screening for malignant neoplasms, colon \\ 
        327.23 & 2.76E-06 & 0.014258581 & * & 0.14363 & 0.10412 & 1.379 & Obstructive sleep apnea (adult) (pediatric) \\ 
        V82.81 & 2.96E-06 & 0.015304732 & * & 0.05346 & 0.02928 & 1.825 & Screening for osteoporosis \\ 
        799.55 & 1.19E-05 & 0.061544119 & ~ & 0.03429 & 0.01627 & 2.108 & Frontal lobe and executive function deficit \\ 
        V03.82 & 1.22E-05 & 0.062936385 & ~ & 0.05724 & 0.03362 & 1.702 & Need for Streptococcus pneumoniae vaccination \\ \bottomrule \\ 
    \end{tabular}
    \caption{Top 20 enriched codes for other neuro cluster in notes clustering.}
    \label{table:notes_other_neuro}
\end{table}

\end{landscape}

\newpage

\subsection*{Cluster Comparison}

\begin{figure}[H]
    \centering
    \includegraphics[width=\textwidth]{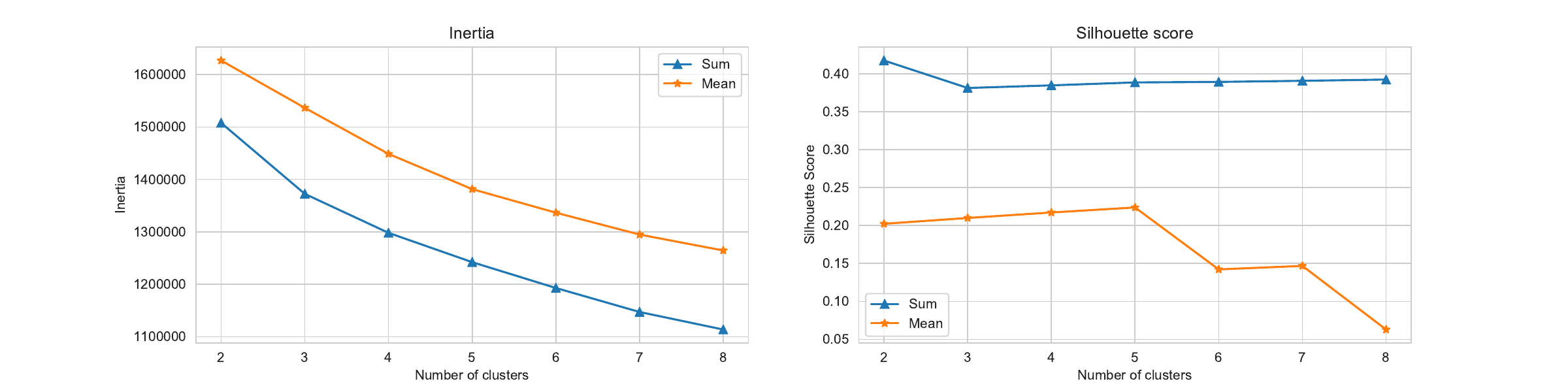}
    \caption{Inertia (within-cluster sum of squares, left) and silhouette score (right) and as a function of number of clusters and embedding methodology for the ICD clustering.}
    \label{fig:cluster_comparison}
\end{figure}

\begin{figure}[H]
    \centering
    \includegraphics[width=\textwidth]{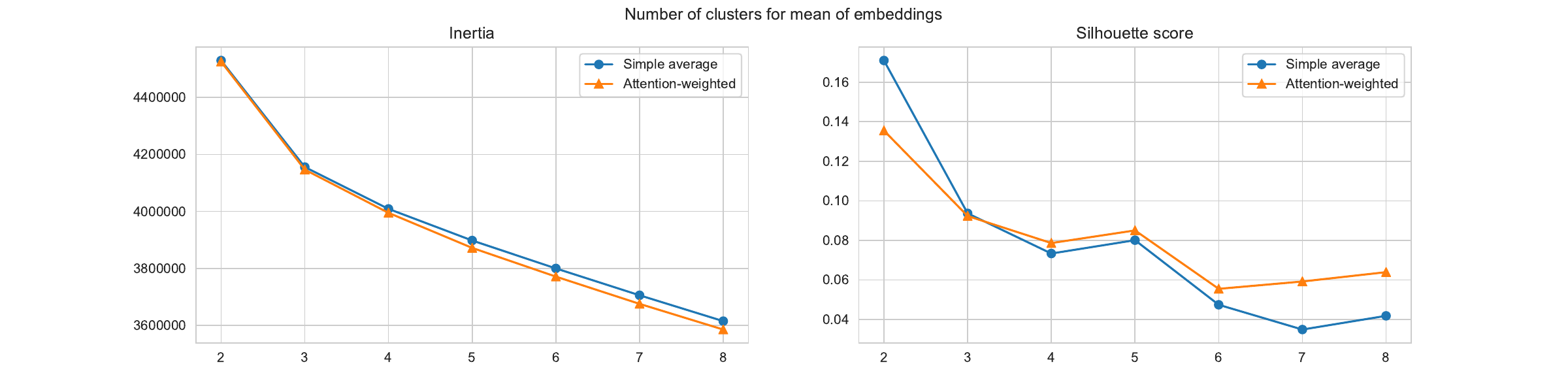}
    \caption{Inertia (within-cluster sum of squares, left) and silhouette score (right) as a function of number of clusters and embedding methodology for the patient notes clustering.}
    \label{fig:notes_cluster_comparison}
\end{figure}

\subsection*{Attention Heatmaps}

\begin{figure}[H]
    \centering
    \includegraphics[width=0.7\textwidth]{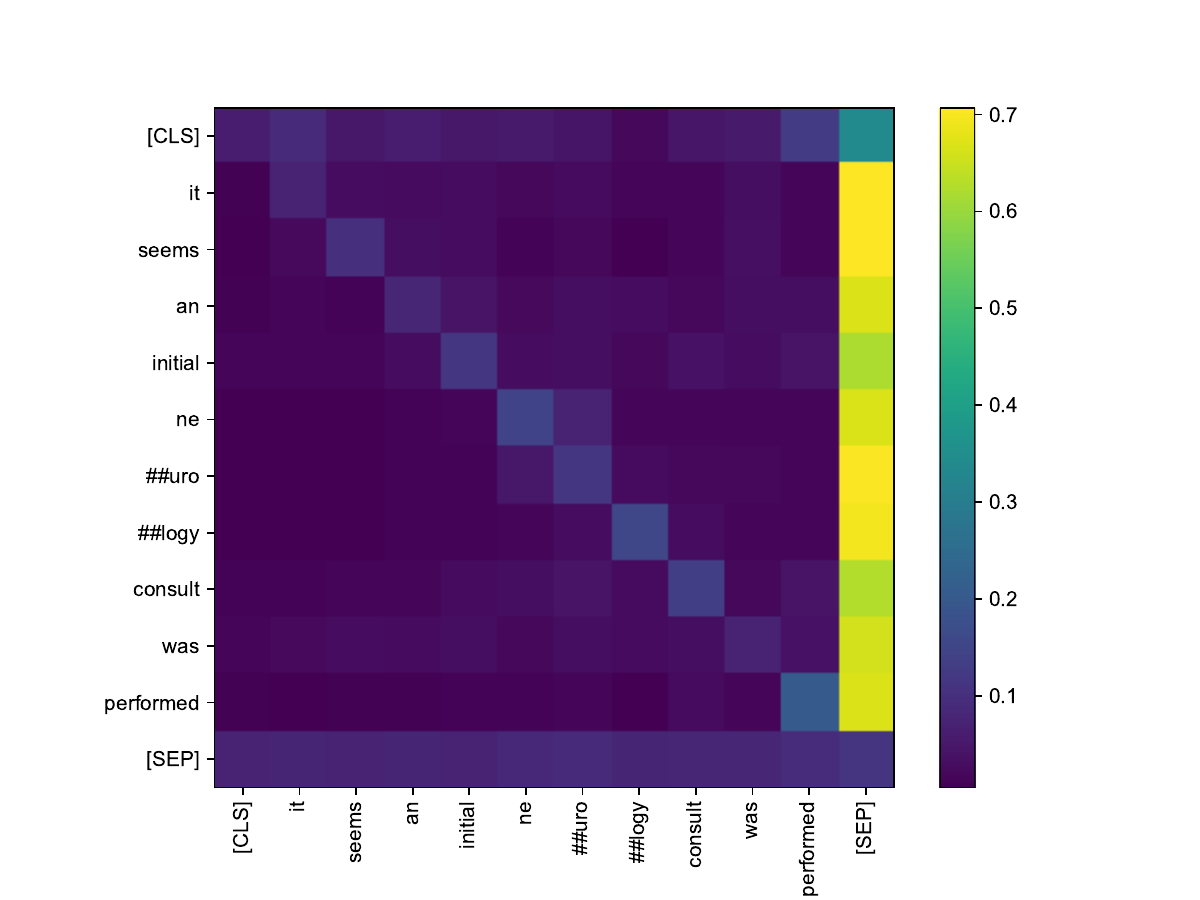}
    \caption{Attention heatmap of final layer for example input sequence, averaged over all 12 attention heads. This attention matrix exhibits high attention weights to the [SEP] token for most tokens, and slightly higher row-wise entropy (and therefore higher weight in averaging) for `initial' and `consult'.}
    \label{fig:attention heatmap_1} 
\end{figure}

\begin{figure}[H]
    \centering
    \includegraphics[width=0.7\textwidth]{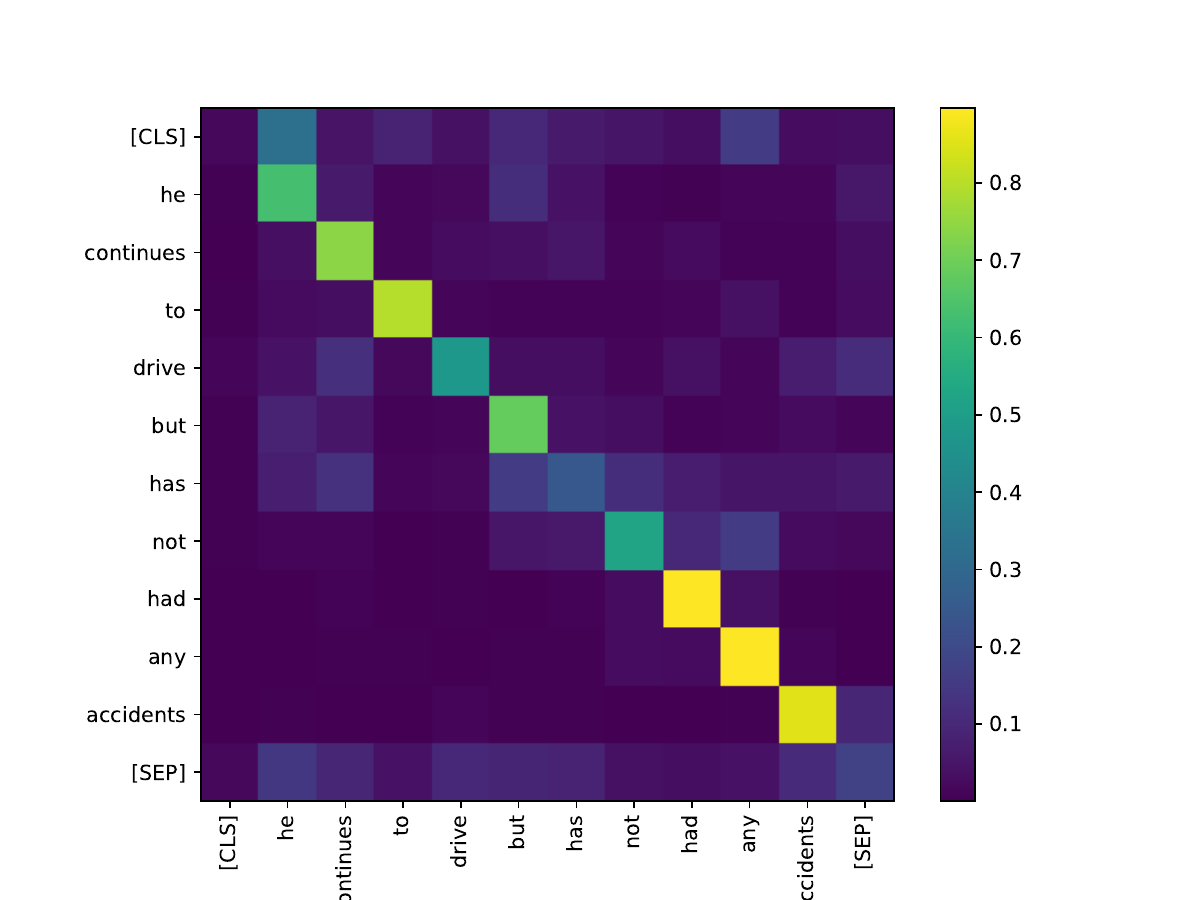}
    \caption{Attention heatmap of final layer for example input sequence, for head 3 alone. Most tokens attend heavily to themselves, indicating relatively low row-wise entropy. Examples of tokens with higher-row wise entropy that attend to multiple tokens include `drive', `has', and `not'.}
    \label{fig:attention heatmap_2}
\end{figure}